\documentclass[runningheads]{llncs}
\usepackage{graphicx}

\usepackage[caption=false,font=footnotesize]{subfig}
\let\MYorigsubfloat\subfloat
\renewcommand{\subfloat}[2][\relax]{\MYorigsubfloat[]{#2}}

\usepackage[utf8]{inputenc}
\usepackage[export]{adjustbox}

\usepackage{makecell}
\usepackage{amssymb}
\usepackage{graphicx}

\usepackage[T1]{fontenc}    % use 8-bit T1 fonts
\usepackage{hyperref}       % hyperlinks
\usepackage{url}            % simple URL typesetting
\usepackage{booktabs}       % professional-quality tables
\usepackage{amsfonts}       % blackboard math symbols
\usepackage{nicefrac}       % compact symbols for 1/2, etc.
\usepackage{microtype}      % microtypography
\usepackage{xcolor}         % colours
\usepackage{algorithm2e}
\usepackage{amsmath}
\usepackage{wrapfig}

\begin{document}
\title{When Neural Networks Using Different Sensors Create Similar Features}
\author{
Hugues Moreau  \inst{1,2}\orcidID{0000-0002-0569-4190}  \and
Andréa Vassilev \inst{1} \and
Liming Chen   \inst{2}\orcidID{0000-0002-3654-9498}}

\authorrunning{H. Moreau et al.}
\institute{Université Grenoble Alpes, CEA, Leti, F-38000 Grenoble,  France  \email{name.surname@cea.fr } \and
Department of Mathematics and Computer Science, Ecole Centrale de Lyon, University of Lyon, Ecully, France \email{name.surname@ec-lyon.fr}}

\maketitle              % typeset the header of the contribution
\begin{abstract}
Multimodal problems are omnipresent in the real world: autonomous driving, robotic grasping, scene understanding, etc... 
Instead of proposing to improve an existing method or algorithm: we will use existing statistical methods to understand the features in already-existing neural networks. 
More precisely, we demonstrate that a fusion method relying on Canonical Correlation Analysis on features extracted from Deep Neural Networks using different sensors is equivalent to looking at the output of the networks themselves. 
%More precisely, we demonstrate that for each sensor, the linear combination of the features from the last layer that correlates the most with other sensors corresponds to the classification components of the classification layer. We will use this to show that the features learnt by a pair of neural networks are extremely similar, even when the neural networks use different sensors. 
\keywords{Multimodal Sensors \and Deep Learning \and Transport Mode Detection \and Inertial sensors \and Canonical Correlation Analysis}
\end{abstract}

\section{Introduction}

Picture a rural scenery: in the countryside, the wind blows through a batch of trees. One can imagine hearing the sound of the wind in the leaves, seeing the branches bend to the gusts of wind, or even feeling the cold air on their skin. All of these stimuli are linked to a single event. Our world is multi-modal: at all times, any event can be captured using a broad diversity of channels. Many real-life problems rely on using multiple modalities: vision and LIDAR sensors for autonomous driving, visual and haptic feedback for robotic grasping, humans even use multiple modalities to understand each other, reading on the lips of their interlocutors.

In the Machine Learning community, a great deal of literature exists to leverage multiple sensors. Some publications use problem-specific solutions, but some approaches are generic: one can, for instance, give the information from all sensors to a single neural network. Or, one can choose to create one network per sensor, to train them to solve the problem the best they can, and to merge the predictions afterwards. In particular, Ahmad \textit{et al.} and Imran \textit{et al.} (\cite{ahmad_human_2020,imran_evaluating_2020} respectively) performed the fusion using a statistical tool named Canonical Correlation Analysis (CCA), in order to find correlations within two sets of features produced by neural networks trained separately. Their goal was to create a new common representation from all sensors for a gesture recognition problem.

The CCA operation has been used in multiple publications to understand deep neural networks working on a    single-modality problem, \cite{raghu_svcca:_2017,morcos_insights_2018,kornblith_similarity_2019}. In particular, Roeder \textit{et al.} \cite{roeder_linear_2020} demonstrated that several architectures, using the same input data, are approximately equal up to a linear transformation. This impressive result was soon followed by McNeely-White \textit{et al.} \cite{mcneely-white_exploring_2021}, who showed a similar result for networks working on face recognition. 
%This impressive result was soon followed by McNeely-White \textit{et al.} \cite{mcneely-white_exploring_2021}, who showed the most correlated components obtained with CCA are equal to the components of the following classification layer for a single sensor. 

The present work extends this claim and helps to understand the similarity between the feature neural networks learnt from different sensors. More precisely, we show that the most correlated components between the features from different \textit{sensors} are equal to the class components, \textit{i.e.}, the vectors forming the column of the weight matrix from the classification layer. The most short-term consequence is that the fusion method introduced in \cite{ahmad_human_2020,imran_evaluating_2020} is equivalent to an average of predictions. 
To sum up, our contributions are the following: 
\begin{itemize}
    \item we demonstrate that the CCA recomputes the information from the classification layer of the network
    \item we apply this reasoning to show the fusion methods introduced in \cite{ahmad_human_2020,imran_evaluating_2020} is identical to a mere average of class logits.
\end{itemize}

We want to emphasize that we use existing methods and algorithms to reach a new conclusion, which is to show that the use of CCA for data fusion can be replaced by a much less complex equivalent. 
The rest of this work is organized as follows: 
section \ref{sec:problem_presenttion} introduces some notations, reviews the Canonical Correlation Analysis, and explains some fundamental concepts to understand our work. Then, we show how the present work is novel compared to the rest of the literature in section \ref{sec:related_works}. Finally, section \ref{sec:experiments} explains the experiments we led and analyzes the results.

\section{Problem position}\label{sec:problem_presenttion}

\subsection{Deep Feature Extraction}\label{subsec:deep_features}

\begin{figure}
    \centering
    \includegraphics[width=4in]{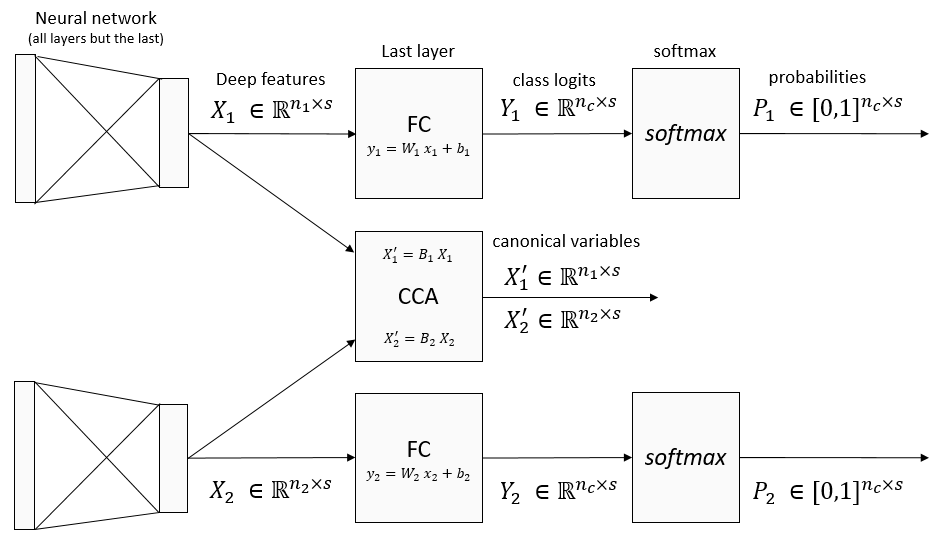}
    \caption{The extraction of deep features. $n_c$ is the number of classes, $n_i$ is the number of features from each feature matrix, and $s$ is the number of samples}
    \label{fig:feature_extraction}
\end{figure}

Let us consider two networks, either two initializations of networks using the same sensor, or networks using different sensors. The most common way to extract features from a network is to record the hidden features right before the last layer (the classification layer), as fig. \ref{fig:feature_extraction} illustrates. We name these feature matrices $X_1$ and $X_2$. An important point to note is that these features are computed from the same samples: if the $i^{th}$ line of $X_1$ is recorded using an accelerometer segment recorded at a given date, the $i^{th}$ line of $X_2$ must be computed from data (for instance, magnetometer data) recorded at the same exact moment as the accelerometer segment. The sensors may differ, but the intrinsic samples (and their order in the feature matrices $X_{1,2}$) must correspond.  

We call $W_i \in \mathbb{R}^{n_c \times n_i}$ (where $n_c$ is the number of classes and $n_i$ is the number of features from each feature matrix) the \textit{class components}, that is, the column vectors of the weights of the last fully-connected layer (the middle layer in fig. \ref{fig:feature_extraction}). As with every other matrix multiplication, one can understand the classification process $x \rightarrow W_i.x$ (we omit the bias) as a series of scalar products with the $n_c$ column vectors of $W_i$: for each class $c$, the scalar product between each feature vector $x$ and the $c^{th}$ column of $W_i$ gives the logit of class $c$, a real number giving the likelihood for the sample to belong in class $c$ (the higher the number, the higher the chances that the sample belongs in the class). These logits are then fed into the softmax operation, in order to obtain a series of probabilities ($n_c$ numbers between 0 and 1 that sum to 1).

\subsection{Canonical Correlation Analysis}\label{subsec:CCA_presentation}

Canonical Correlation Analysis is a statistical tool that takes two feature matrices $X_1$, $X_2$, and returns a series of linear combinations of each of these features $X'1 = B_1.X_1$, $X'_2 = B_2.X_2$ (where $B_1, B_2$ are basis change matrices). These new features are defined recursively: the first column of $X'_1$ and the first column of $X'_2$ (the first \textit{canonical variables}) are computed such that the correlation between them is maximized. Then, the second columns of these matrices maximize the correlation between each other while being decorrelated to the previously computed (the correlation between the first and the second columns of $X'_1$ is zero). The subsequent columns are constructed the same way, by maximising the mutual correlation between matrices $X'_1$ and $X'_2$, while being decorrelated to previously computed components. Note that this requires the matrices $X_i$ to be full-rank so that we have enough components. In practice, we use PCA to obtain full-rank feature matrices (we remove the components that account for less than 0.01 \% of the cumulative variance).

Similarly to the class components, the \textit{canonical components} are the column vectors of the $B_i$, and we will compare the first of them to the class components. There are as many canonical components as there are input features in the feature matrix $X_{1,2}$, but we are only interested in the \textit{first} $n_c$ components. They correspond to the $n_c$ \textit{most} correlated components one can find in the features. We want to show there is a linear relationship between these first $n_c$ canonical components and the $n_c$ class components.

\subsection{A simplistic example}\label{subsec:main_claim}

Let us consider an unrealistic, but illustrative, example, and let us imagine that the class logits were equal across networks $Y_1 = Y_2$. One should notice that the logits are linear combinations of features $Y_i = W_i.X_i$\footnote{in the following sections, we will omit the bias in the equation $Y_i = W_i.X_i +b_i$. As the CCA assumes that $X_{1,2}$ have zero mean, a necessary step prior to the computation of the canonical components is to remove the mean of the features $X_i$. This is why adding the constant bias $b_i$ does not change anything to the reasoning}. This means that one can find linear combinations of features that correlate perfectly with each other. Yet, because of the way the canonical components are computed, these class components will always appear first among the canonical components.
In practice, the logits are not equal, but they only need to be correlated enough to each other. 
 
Once this is understood, it is easier to understand the main claim of this work. If we consider two networks, producing the sets of features $X_1$ and $X_2$, that succeed fairly at the same classification task, then, the logits $Y_1$ and $Y_2$ produced by those networks will be correlated. The last layer of the networks is linear, in other words, the class components can be found among both feature vectors $X_1$ and $X_2$ with a simple linear transformation ($Y_i = W_i.X_i$). This means that if one applies CCA to the couple of feature matrices $(X_1, X_2)$, one can find the class components $W_i$ among the first components of $B_i$.

\begin{figure}
    \centering
    \includegraphics[width=10cm]{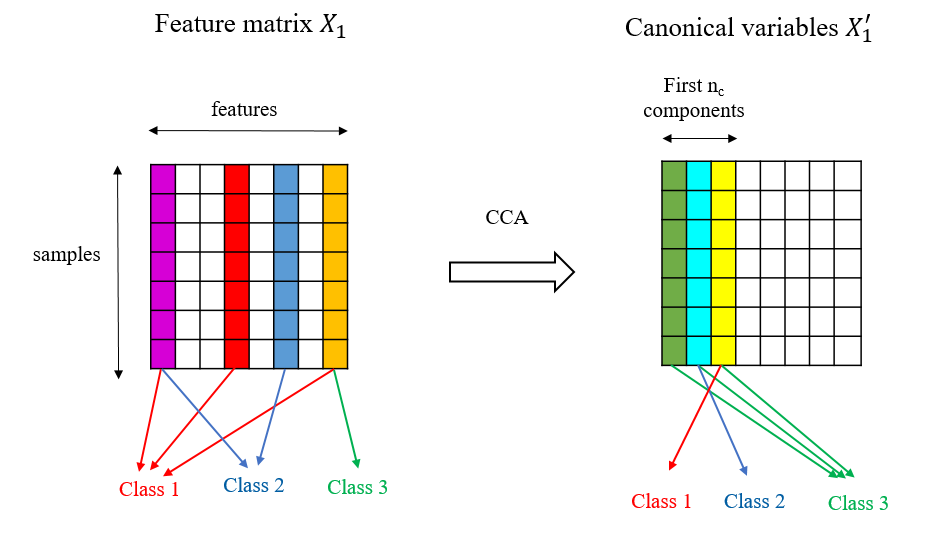}
    \caption{The principle of the equality between class components and the first canonical components on a three-class problem. The colours in the different feature matrices denote the different information about the three classes. The feature vectors will undergo a matrix multiplication (denoted by the arrows under the left matrix); and the rows of the matrix the features are multiplied by are the class components.}
    \label{fig:components_equality_illustration}
\end{figure}

In particular, this means computing the sum of the canonical variables ($X'_1+X'_2$, as \cite{ahmad_human_2020,imran_evaluating_2020} do) is equivalent to summing the logits $Y_1 + Y_2$. Figure \ref{fig:components_equality_illustration} illustrates how the equality of the canonical components and the class components make the CCA fusion equivalent to a sum of the logits. 

\subsection{Extensions}\label{subsec:extensions}

Section \ref{subsec:main_claim_experiment} will detail the experiments we lead to demonstrate the correspondence between CCA components and classification components, both in the case of networks using the same sensor, and in the case of networks using different sensors.

One might wonder what are the causes of the phenomenon. If we sum up the previous sections, for CCA to pick up the class components, two conditions must be verified: 
\begin{itemize}
    \item The class logits of two networks must be more correlated than any other component of the feature space. 
    \item We must apply CCA on the features from the last layer (so that there is a linear relationship between the features and the class logits)
\end{itemize}

The first condition seems to be verified in practice. For instance, \cite{mania_model_2019} studied the prediction similarity of networks working on the same problem, and found that they are much more similar to each other than what their accuracies could lead to believe. %This is a condition we will verify in section \ref{subsec:correlation_logits}. 
As for the second condition, some publications work with CCA in other layers than the last \cite{raghu_svcca:_2017,morcos_insights_2018} to explore the behaviour of neural networks. 
%We will verify empirically if the CCA components still encode the classification information in section \ref{subsec:different_layers}.
However, the verification of this second condition (especially finding an equivalent to the classification components in the earlier layers), is out of scope for this work.

\section{Related Works}\label{sec:related_works}

\subsection{CCA as a fusion method}\label{subsec:CCA_fusion}
Two works \cite{ahmad_human_2020,imran_evaluating_2020} used the Canonical Correlation in a multimodal setting: in a problem with several sensors (each of them being able to bring some information about the problem), they both considered the following process: first, they trained one neural network per sensor. Then, they extracted the hidden representations from the last layer of each network. They computed the canonical variables ($X'_1$ and $X'_2$ with the notations from the previous section), then summed these components ($X'_1+X'_2$), before using a Machine Learning algorithm (SVM or KELM) to guess the final prediction from the result of the addition.

Table 7 from \cite{imran_evaluating_2020} shows the results are not very different from averaging of the output \textit{probabilities} of each network. One of our contributions is to show that the CCA operation isolated the class logits in the first components of $X'_1$ and $X'_2$ and that classifying the sum $X'_1+X'_2$ is roughly equivalent to summing the output \textit{logits} of each network, class by class.

\subsection{Similarity of different neural networks}
The similarity between two neural networks is a well-studied subject. The closest publications to our work are the ones from Roeder \textit{et al.} \cite{roeder_linear_2020}. In 2020, they discovered that the representations learnt by different architectures working using the same input data are equal up to a linear transformation; for a broad diversity of tasks including classification. Since this publication, other works, such as \cite{mcneely-white_exploring_2021}, expanded the claim to a broad diversity of monomodal architectures and brought new experiments validating this information. 

To uncover the similarity between deep models, others looked more at the predictions of networks. For instance, \cite{hacohen_lets_2020} studied the order in which different models learn to classify each sample, while \cite{mania_model_2019} demonstrated that two neural networks classify the samples the same way. 

%All these results indicate that there is a common point, a behaviour, that is shared between architectures. This might be due to the training process: Arora \textit{et al.} \cite{arora_implicit_2019} demonstrated that linear systems composed of multiple layers trained by gradient descent tended to learn "simple" (\textit{i.e.} low-rank) solutions first. 

However, all these studies work on monomodal problems, that is, problems with a single input (in most cases, image classification). In the other hand, we provide an example of the similarity between representations learnt with the same architectures, using different \textit{sensors}. The implications are important: this means the networks learnt to exploit the information that remains common between sensors.

\subsection{SVCCA and improvements}
Several publications worked to improve the computation of the similarity between two bases of features. The first one is SVCCA \cite{raghu_svcca:_2017}, a famous publication that popularized the use of CCA to measure the similarity between two networks. The idea is to use PCA (Singular Vector decomposition, hence the SV in the abbreviation) on each of the feature matrices to remove low-variance components (which are assumed to be noise), before applying CCA on the reduced feature matrices. We too apply PCA, but only to remove the components with negligible variance (we keep $99.99 \%$ of the variance). That is, we keep the 'noisy' low-variance components. To compare, the authors from \cite{raghu_svcca:_2017} keep only $99 \%$ of the variance, that is, they remove 100 times more variance than us. This means the results we draw are more robust. 
Morcos \textit{et al.} \cite{morcos_insights_2018} also noticed the components found by the classic CCA could be noisy. When measuring the proximity between two sets of features $X_1$ and $X_2$, they still compute the CCA components $X_{1,2}'$, but instead of using the average of the correlations between $X'_1$ and $X_2'$, they choose one of the feature sets (let us say, $X_1$), and weight the correlation proportionally to the variance which is kept by each CCA component (\textit{i. e.}, the variance of each component of $X_1'$ divided by the total variance of $X_1$). This method, named Projection-Weighted CCA (PWCCA), is better at rejecting noise than SVCCA. 

Finally, Kornblith \textit{et al.} \cite{kornblith_similarity_2019} extended upon this approach, by dividing the correlation by the relative variance of both bases ($X_1'$ and  $X_2'$). They named their method CKA (Centered Kernel Alignment) because they use the kernel trick to find better alignments than mere linear combinations. 
%Note that the formula for the linear version of CKA is the opposite of the Grassmann distance \cite{hamm_grassmann_2008} between the feature matrices $X_1$ and $X_2$, seen as subspaces of $\mathbb{R}^{s}$ (where $s$ is the number of samples).

As \cite{kornblith_similarity_2019} states, these two methods are closely related. PWCCA \cite{morcos_insights_2018} consists in re-weighting the CCA components by the variance of one base, while linear CKA consists in re-weighing the components by using both variances (relatively to the variance of the original sets $X_{1,2}$). To summarize, one can see PWCCA and CKA as different mixtures between PCA and CCA. One could wonder if the conclusions we drew here also apply in the case of PWCCA and CKA. We argue that this is the case, for the following reason: Kamoi \textit{et al.} \cite{kamoi_why_2020} showed that when a network deals with inliers (non-outliers), the components with the highest variance among the features are approximately equal to a combination of the class logits. This explains the high results of the 'PCA' curve in section \ref{subsec:experiment_proj_initialization}. We showed that the most correlated components are very similar to logits. As a consequence, we expect re-weighting the importance of the CCA components by the amount of variance accounted for by each component to enforce even further the proximity between class logits and most important components. 
However, this paper focuses on regular CCA, which means that the experiments extending our conclusions to kernel-CCA or CKA are out of scope. 

%However, exact experiments demonstrating this fact are out of scope for this work, as for the extension of our conclusions to kernel-CKA. We will focus on the most basic CCA, as the following section explains. 

\section{Experiments}\label{sec:experiments}
In this section, we will first reproduce the results from \cite{raghu_svcca:_2017}, in order to illustrate our experimental protocol (section \ref{subsec:experiment_proj_initialization}). Then, we will repeat this experiment mixing data from different sensors to show our main claim in section \ref{subsec:experiment_proj_sensor}. %Finally, we will try to explore the limits of the phenomenon in section \ref{subsec:limits_experimental}. 

\subsection{Datasets}\label{sec:datasets}

\subsubsection{CIFAR10}
We use the famous ResNet-56 network \cite{he_deep_2016} on the CIFAR-10 Dataset. This is a Computer Vision classification problem, where the model has to classify low-resolution ($32 \times 32$) images into ten classes. The dataset contains 50,000 training samples and 10,000 validation samples. We trained the network hyperparameters and architecture as the original publication \cite{he_deep_2016} thanks to the code from \cite{noauthor_322_nodate}.

The dataset has only one sensor (the RGB images), but we work with different initializations of networks that use the same modality. We use this dataset to provide a comparison with the rest of the literature on CCA with deep features, as most works chose to include a ResNet trained on CIFAR-10 \cite{raghu_svcca:_2017,morcos_insights_2018,kornblith_similarity_2019}.

\subsubsection{SHL 2018 dataset}

The Sussex-Huawei Locomotion 2018 dataset is a Transport Mode detection problem. Organizers asked three participants to record the sensor values from several smartphones while travelling using different modes (walking, running, driving, \textit{etc.}). Then, the data is published, and a yearly challenge is organized to get a precise evaluation of the state of the art. The 2018 dataset is the first version of the challenge: only the data from a single user and a single smartphone (the one in the hand) is available for classification. The organizers released 16,310 annotated samples for training and validation. 

\begin{table}[]
    \centering
    \begin{tabular}{c c}
        \hline
        sensors & \thead{ Accelerometer, Gravity, Linear Acceleration, Gyrometer, \\ Magnetometer, Orientation quaternion, barometric Pressure} \\ \hline
        %Acc, Gra, LAcc, Gyr, Mag, Ori, Pressure \\ \hline
        classes & \thead{Still, Walk, Run, Bike, Car, Bus, Train, Subway} \\ \hline
        segment duration & $60s$ \\ \hline
        sampling frequency & $100 Hz$ \\ \hline
        training samples & $13,000$ \\ \hline
        validation samples & $3,310$ \\ \hline
    \end{tabular}
    \caption{An overview of the SHL 2018 dataset}
    \label{tab:SHL_overview}
\end{table}

The dataset includes seven sensors (accelerometer, gravity, linear acceleration, gyrometer, magnetometer, orientation vector, barometric pressure), most of them having several axes ($x$,$y$,$z$). We will study three signals among them: the $y$ axis of the gyrometer (\texttt{Gyr\_y}), the norm of the acceleration (\texttt{Acc\_norm}, as in \cite{ito_application_2018}), and the norm of the magnetometer (\texttt{Mag\_norm}). The accelerometer and gyrometer encode similar information (they record the inertial dynamics of the sensor) and are most useful when detecting walk, run, or bike segments. On the other hand, the norm of the magnetometer mostly changes when the sensor is close to a strong magnetic field: far from any ferromagnetic object, its values stays close to $40\mu T$ (the value of the Earth's magnetic field). But this sensor can go up to $200 \mu T$ when a strong magnetic field is present (for instance, when the sensor is close to a ferromagnetic object or even an electrical engine). This is why we think this sensor will be best to detect the train or subway classes from the rest.
To summarize, the accelerometer and gyrometer are expected to be similar to each other, while these sensors encode different information than the magnetometer. This is intended to represent different relatedness between sensors.

We use the same approach as in \cite{ito_application_2018}, each signal is first converted into a two-dimensional spectrogram (a time-frequency diagram) using short term Fourier transform. The frequency axis of the spectrogram is rescaled using a logarithmic scale, in order to give more resolution to the lower frequencies. This method aims to give better resolution to the $2-3 Hz$ frequency bands (which are the most useful to distinguish the Walk, Run, and Bike segments), while still keeping the highest frequencies available. See \cite{ito_application_2018} for more information and illustrations. For each sensor, we obtain a $48 \times 48 \times 1$ spectrogram, that is fed into a CNN which architecture is simple: three convolutional layers (with 16, 32, and 64 filters), and two fully-connected layers (with 128 hidden features and 8 output features). See \cite{ito_application_2018} for details about hyperparameter or training process.

To illustrate, on three random initializations, the average validation F1-score of each of these individual sensors is 89 \% for the accelerometer, 80 \% for the gyrometer, and 67 \% for the magnetometer.

In both experiments, we will use a train set to train the models, extract the features, compute the base change with CCA, and, when applicable, retrain the models. The validation sets only go through trained models and already-computed base changes, before being used to display a result. We want to emphasize that when dealing with multimodal sensors, each network was only trained on a single modality: the network using the accelerometer never saw the gyrometer or magnetometer data, and so on.

\subsection{Studying component similarity with subspace projection} \label{subsec:main_claim_experiment}
\begin{figure}
    \centering
    \includegraphics[width=4in]{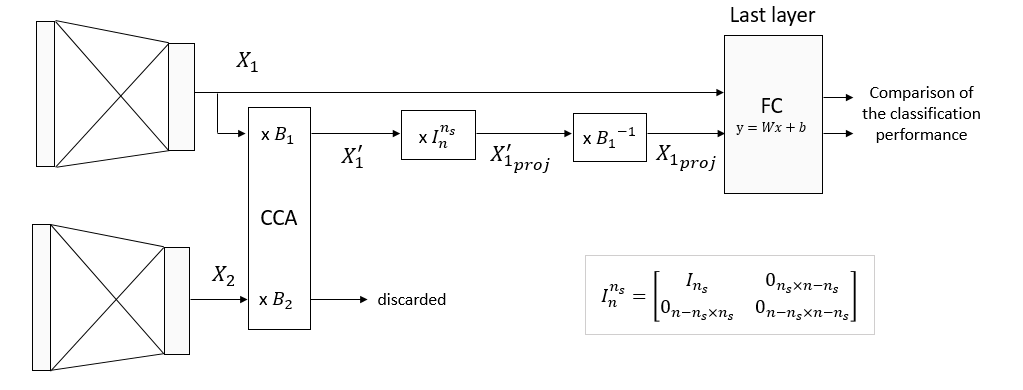}
    \caption{The principle of the subspace projection experiment: $P_1 = B_1^{-1}.I_{n}^{n_s}.B_1$ projects $X_1$ onto a linear space with dimension $n_s$.}
    \label{fig:projecton_experiment}
\end{figure}

In this section, we will reproduce and extend the experiments from \cite{raghu_svcca:_2017} (Figure 2 from this work). As fig. \ref{fig:projecton_experiment} illustrates, we start from a trained network, we extract the hidden features form the last layer, then we project on a subspace of inferior dimension, before re-injecting the features in the network to measure the performance. If the performance is intact, it means that the $n_c$ class components are unaffected by the projection. In other words, it means the class components already belong in the image of the projection. In particular, when the dimension of the image of the projection is $n_s=n_c$, and if the performance is unchanged, it means that the $n_c$ class components belong in the subspace spanned by the $n_s$ most correlated components, which implies the existence of a linear relationship between the families (as the canonical components and class components are both linearly independent families of vectors).

Note that when we use all features, we project on the original space, \textit{i.e.}, we leave the data unchanged. The difference between the end of the curves (performance on pristine data) and the rest (altered data) will indicate the proximity between the considered subspace and class components.

The dimension and the way of choosing the subspace will vary: as in \cite{raghu_svcca:_2017}, we consider choosing the $n$ most correlated components found with CCA (\texttt{CCA\_highest}), the $n$ features with the highest activation in absolute value (\texttt{max\_activation}), and $n$ features chosen randomly (\texttt{random\_selection}). In order to provide comparisons, we add four reductions methods that are not included in \cite{raghu_svcca:_2017}: 
\begin{itemize}
    \item \textit{random orthogonal projection} (\texttt{random\_projection}). Comparing the random selection of \textit{n} components versus the projection on \textit{n} components shows that the canonical basis does not play a particular role (\textit{i.e.} selecting the values of $n$ features is not particularly meaningful). 
    \item \textit{PCA}: Kamoi \textit{et al.} \cite{kamoi_why_2020} showed that the $n_c$ components with the most variance are the components that will be used for classification. We project the features on the components with most variance to validate their findings.
    \item \textit{least correlated components} (\texttt{CCA\_lowest}): if the most correlated components are the class components, the components with lowest correlation should not include any relevant information for the problem.
    \item \textit{CCA with random components} (\texttt{CCA\_random}): one may argue that the CCA curve is above the others in \cite{raghu_svcca:_2017} because CCA allows to create decorrelated components, which would mean that its components are less redundant than random directions. If this was the case, selecting random CCA components would be better than selecting components with a random projection. 
\end{itemize}

To save time, we do not consider all the possible number of components: because we want a high resolution around $n_c$, we only considered the $2*n_c$ first components (where $n_c$ is the number of classes, 8 for SHL and 10 for CIFAR), and, after that, the number of components which are powers of 2 ($16, 32, ...$), up to the maximal number of components ($128$ for SHL, $64$ for CIFAR)

In addition to this, after measuring the performance of the layer when using projected features, we also try retraining the classification layer on a projected version of the validation set, with the same hyperparameters as the initial training of the network. The goal of this retraining is to illustrate the difference between the components a network actually uses for classification and the components that carry some information about the problem. If the performance of the retrained layer is low, this means we can be sure that the projection removed \textit{all} useful information. If only the performance of the original layer is low, this only means that we got rid of the information that was used by the network.

Note that the CCA operation requires two databases. When we use CCA, we use a second matrix of features $X_2$, but only to compute $X'_1$ (we discard $X'_2$). In the next section, this second network is another initialization of a network working with the same sensor, while the section after that shows experiments made with two networks using different sensors.

\subsubsection{Similarity between identical sensors} \label{subsec:experiment_proj_initialization}

\begin{figure*}
\centering
\subfloat[CCA_monomodal_CIFAR]{\includegraphics[height=2.5in]{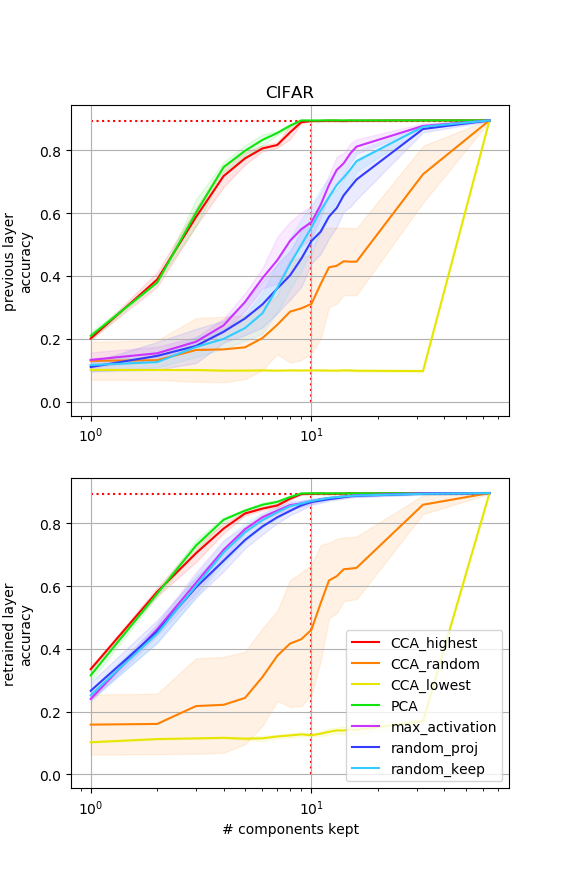}
  \label{fig:CCA_monomodal:CIFAR}}
\subfloat[CCA_monomodal_SHL]{\includegraphics[height=2.5in]{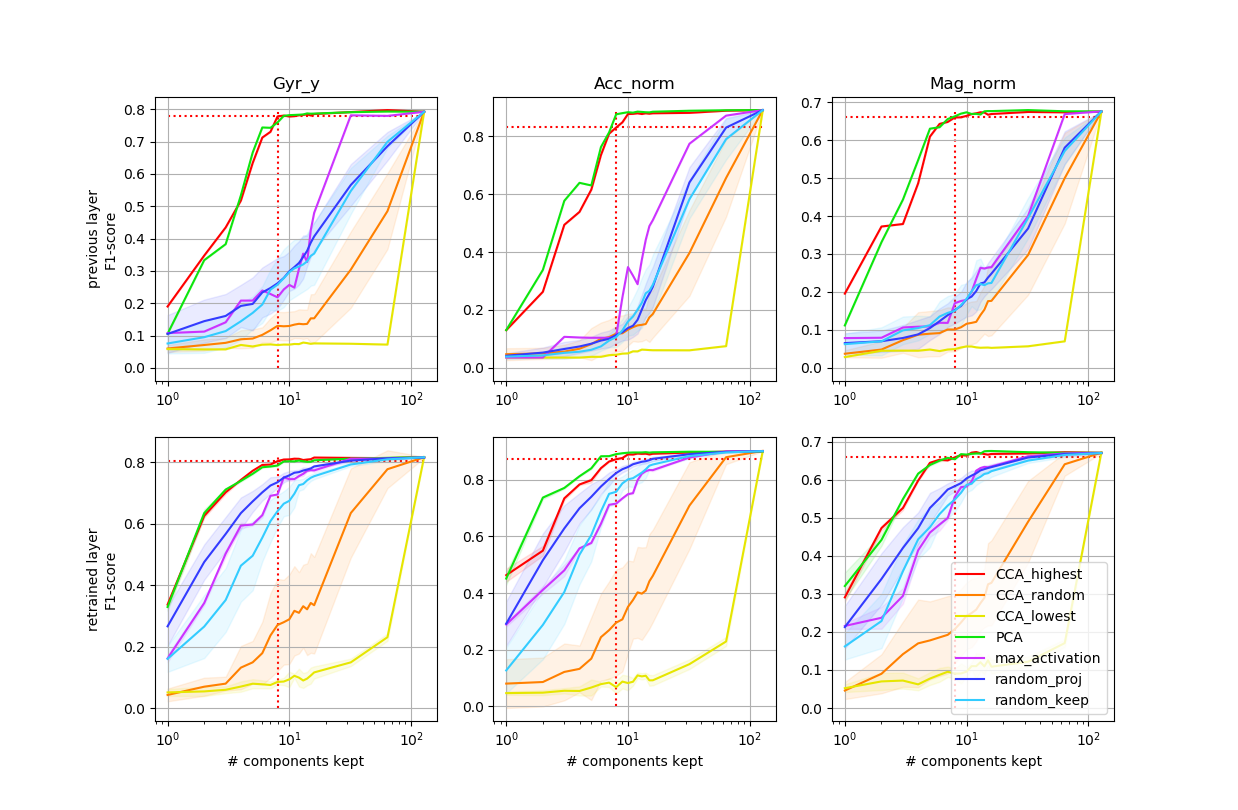}
  \label{fig:CCA_monomodal:SHL}}
\caption{The performance of the networks after projecting their features on subspaces with varying dimensions, on the CIFAR (a) and SHL (b) validation sets. The top row indicates the validation performance of the network as is, while the bottom row indicates the performance when retraining the classification layer on a projected training set. For each curve, the experiment was repeated 5 times, and the standard deviation is given by the width of the curve (which is sometimes too small to see). The dotted line highlights the performance with the $n_c$ most correlated components. Best view in colour.}
\label{fig:CCA_monomodal}
\end{figure*}

Figure \ref{fig:CCA_monomodal} shows the result of this experiment. We can draw several conclusions from it:
\begin{itemize}
    \item The performance of the projection on the $n_s$ highest variance components ('pca', green curve) is maximal for $n_s=n_c$: this verifies the findings of Kamoi \textit{et al.} \cite{kamoi_why_2020}, the $n_c$ components with highest variance are the class components.
    
    \item Similarly to Figure 2 from \cite{raghu_svcca:_2017}, the most correlated components are more useful for the classification problem than a random choice of components from the canonical basis. 
    
    \item The red curve (performance of the components with the highest correlation) is almost at its maximum for $n_c$ components even before retraining, there is almost nothing to gain after $n_c$ components. This validates our original claim these components correspond to the subspace used by the classification layer. 
    
    \item The yellow curve (the components with lowest correlations), is under all the others. Before retraining, the performance of a projection on the $n$ components with the lowest correlation is minimal, even when we select half the components. After retraining, the performance of these components is still well under the performance of random components: selecting the least correlated components effectively removed most of the classification information. 
    
    \item The orange curve (random CCA components) is lower than the random choice of components (blue curves). This means that the performance of the components with the highest correlation is not due to an efficient encoding. Additionally, the standard deviation of this curve is unusually high: as the CCA operation isolates the classification components from the rest, selecting some of its components at random creates extremes situations: either a classification component is selected, or it is not. The standard deviations of the other random methods are not as high because the random choices allow to span partly the classification components. 
    
    \item Before retraining, the two blue curves are equivalent, this indicates that the canonical components do not play any specific role in regards to classification. After retraining, the dark blue curve (random projection) is higher than light blue (random selection of canonical components). We hypothesize that the canonical components carry some redundancy between them because of the dropout we used to train our networks, and that re-training the network allows it to stop expecting this redundancy in the features it sees. 

\end{itemize}

\subsubsection{Similarity between two different sensors} \label{subsec:experiment_proj_sensor}

We now lead experiments to verify our main contribution: the fact that the most correlated features are equal to classification components, even when the correlation is computed across sensors. This time, when computing CCA, we use features from a network using different sensors. In this section, we do not include any of the other dimensionality reduction methods (PCA, random projection and selection of components, maximal activation components) because those methods work with only one database: the results would be copies of the curves presented in fig. \ref{fig:CCA_monomodal}. 

\begin{figure*}
\centering
\includegraphics[width=5in]{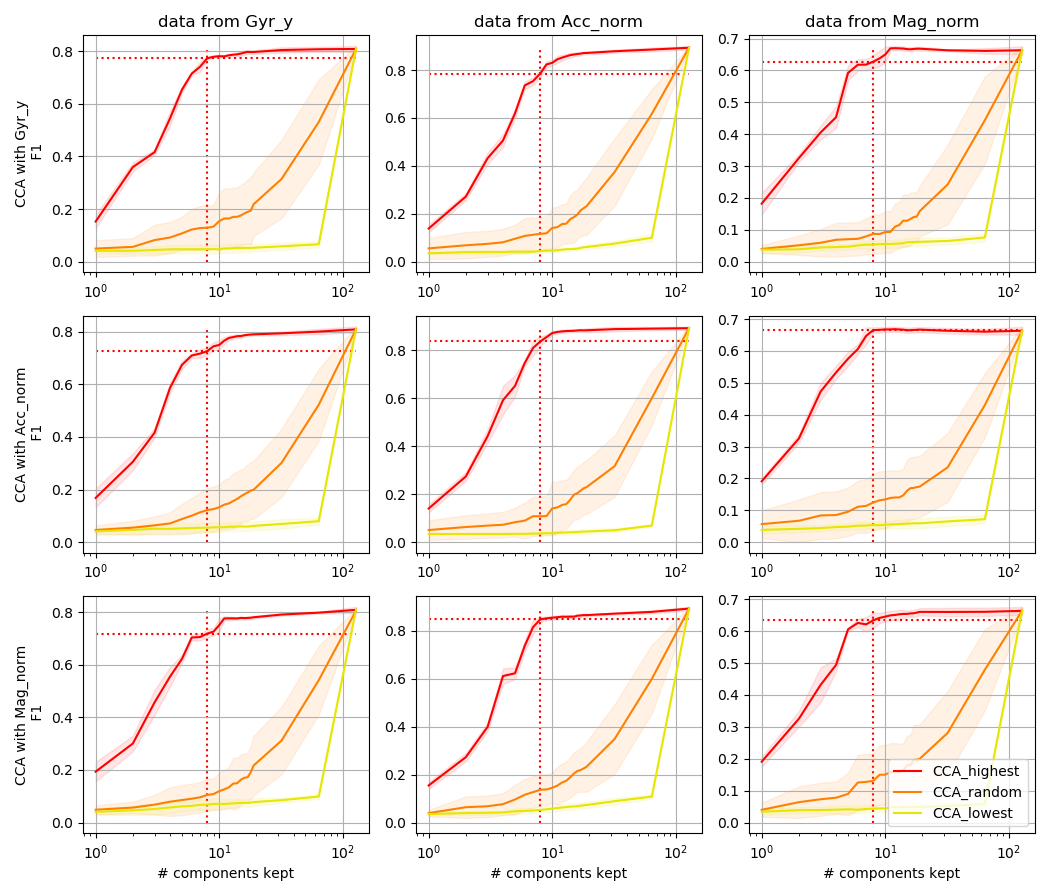}
\caption{The classification performance of classification layer using features projected on a subspace with varying dimension, when the CCA is computed thanks to data from another sensor. As in fig. \ref{fig:CCA_monomodal}, one can see that the performance with the $n_c$ most correlated components is close to the performance with all components. The graphs in the diagonal were generated using the same protocol as the first row of graphs in fig. \ref{fig:CCA_monomodal:SHL}. The dotted line highlights the performance with the $n_c$ most correlated components. Best view in colour.}
\label{fig:CCA_multimodal}
\end{figure*}

Figure \ref{fig:CCA_multimodal} shows that the performance is maximal when the number of components is close to 10 or 20, approximately. However, contrary to fig. \ref{fig:CCA_monomodal:SHL}, the performance is off by a few points when the number of selected components is equal to 8, the number of classes. This means that the equality between most correlated components and classification vectors is less strong than in the previous case when the CCA was computed from the same sensor. Still, the performance with only 8 components is high enough for us to conclude that the components computed with CCA overlap significantly with the classification components.

\section{Conclusion}

We began by demonstrating the experiments from previous publications: the findings from Kamoi \textit{et al.} \cite{kamoi_why_2020}, who showed that the components with most variance are the classification components, and Roeder \textit{et al.} \cite{roeder_linear_2020}, who showed that CCA found the classification components when it is applied to features from monomodal networks using the same dataset. In a later section, we showed the same result held when applied to features learnt from different sensors, indicating that the networks exploit information that can be found across multiple sensors. The exact nature of this information, however, is yet to be found. 

%Finally, we showed that this phenomenon is proper to the features from the last layer of networks and that the non-linearities in the multiple layers prevented the CCA to find components that are the inverse image of the class logits. 
In addition to showing that the fusion method from \cite{ahmad_human_2020,imran_evaluating_2020} is unnecessarily complex, these results have strong implications for multimodal learning: in this situation, it may be unnecessary to add too many sensors, for neural networks would compute similar information.

Future work might include finding exactly the nature of the common information which is present in all sensors' signals and exploited by neural networks. Or, we could try to explain a paradox about the similarity we measured: the networks using the accelerometer, gyrometer, and magnetometer have average F1-scores of about $90\%$,  $80\%$, and  $67\%$ (respectively). How can couples of features that are so similar have such different performance levels?

\bibliographystyle{splncs04}
\bibliography{references}

\end{document}